\documentclass[manuscript]{acmart}

\usepackage{multirow}

\AtBeginDocument{%
  }

\copyrightyear{2024}
\acmYear{2024}
\setcopyright{rightsretained}
\acmConference[FAccT '24]{The 2024 ACM Conference on Fairness, Accountability, and Transparency}{June 3--6, 2024}{Rio de Janeiro, Brazil}
\acmBooktitle{The 2024 ACM Conference on Fairness, Accountability, and Transparency (FAccT '24), June 3--6, 2024, Rio de Janeiro, Brazil}\acmDOI{10.1145/3630106.3658937}
\acmISBN{979-8-4007-0450-5/24/06}

\begin{document}

\title[Gender Bias Detection in Court Decisions]{Gender Bias Detection in Court Decisions: A Brazilian Case Study}

\author{Raysa Benatti}
\email{raysa.benatti@uni-tuebingen.de}
\affiliation{%
  \institution{University of Tübingen}
  \country{Germany}
}
\affiliation{%
  \institution{Universidade Estadual de Campinas (UNICAMP)}
  \country{Brazil}
}

\author{Fabiana Severi}
\email{fabianaseveri@usp.br}
\affiliation{%
  \institution{Faculdade de Direito de Ribeirão Preto, Universidade de São Paulo (USP)}
  \country{Brazil}
}

\author{Sandra Avila}
\email{sandra@ic.unicamp.br}
\affiliation{%
  \institution{Instituto de Computação, Universidade Estadual de Campinas (UNICAMP)}
  \country{Brazil}
}

\author{Esther Luna Colombini}
\email{esther@ic.unicamp.br}
\affiliation{%
  \institution{Instituto de Computação, Universidade Estadual de Campinas (UNICAMP)}
  \country{Brazil}
}

\begin{abstract}
Data derived from the realm of the social sciences is often produced in digital text form, which motivates its use as a source for natural language processing methods. Researchers and practitioners have developed and relied on artificial intelligence techniques to collect, process, and analyze documents in the legal field, especially for tasks such as text summarization and classification. While increasing procedural efficiency is often the primary motivation behind natural language processing in the field, several works have proposed solutions for human rights-related issues, such as assessment of public policy and institutional social settings. One such issue is the presence of gender biases in court decisions, which has been largely studied in social sciences fields; biased institutional responses to gender-based violence are a violation of international human rights dispositions since they prevent gender minorities from accessing rights and hamper their dignity. Natural language processing-based approaches can help detect these biases on a larger scale. Still, the development and use of such tools require researchers and practitioners to be mindful of legal and ethical aspects concerning data sharing and use, reproducibility, domain expertise, and value-charged choices. In this work, we (a) present an experimental framework developed to automatically detect gender biases in court decisions issued in Brazilian Portuguese and (b) describe and elaborate on features we identify to be critical in such a technology, given its proposed use as a support tool for research and assessment of court~activity. 
\end{abstract}

\ccsdesc[500]{Applied computing~Law}
\ccsdesc[500]{Computing methodologies~Information extraction}
\ccsdesc[500]{Computing methodologies~Supervised learning by classification}

\keywords{gender bias, natural language processing, social computing, legal text}

\received{20 February 2007}
\received[revised]{12 March 2009}
\received[accepted]{5 June 2009}

\maketitle

\section{Introduction}
\label{sec:intro}

Natural language processing (NLP) techniques have been proposed to address issues in many domains. Specific fields, such as the social sciences, are more prone to use and produce texts containing relevant data for analysis. The legal field, in particular, has been the focus of interest of many practitioners and researchers who propose techniques to perform tasks such as document classification~\cite{Undavia2018, Silva2018}, information extraction~\cite{pereira24}, and text summarization~\cite{Kanapala2019, Merchant2018}. 

The increase of digital data availability in such domains plays a significant role in using NLP to address some of their inquiries. In recent decades, public institutions have replaced physical documents and procedures with digital ones in many jurisdictions. We stress the Brazilian case: being the most populated Latin American country, it has a substantial court system\footnote{The country has one lawyer for each batch of around 150 people \cite{oab} and approximately 80 million active legal cases \cite{br_jus_yearbook}.}, with large databases of judicial documents and an engaged community focused on developing computational approaches for the legal field. In that sense, it has emerged as a legal data hotspot.

Although increasing procedural efficiency is the primary motivation behind most artificial intelligence-based solutions in legal systems, they can also be explored to address other issues. The possibility of analyzing content on a larger scale offers new methods of investigation and expands the range of research questions about social institutions to be explored. In that context, NLP can be framed as a support tool for assessing court activity.   

One aspect that might be extracted from court decisions raises concerns: the presence of gender biases or stereotypes encrusted in legal reasoning, especially in cases of gender-based violence. There is evidence that court rulings can bear those biases, and NLP approaches can help detect them on a larger scale; however, despite their technical promises and accomplishments, legal and ethical considerations must be carried out when designing, developing, and using such approaches. 

As a case study to support this argument, we introduce an experimental framework of data construction and text classification designed to automatically detect gender biases in court decisions issued in Brazilian Portuguese in the context of gender-based violence-related cases. The pipeline includes structuring data and metadata and developing an attention-based deep-learning solution for its classification. Therefore, it provides a methodological possibility for domain experts to find new answers to their inquiries. We describe the methodology used for developing the framework, the experiments and their main results, and a baseline evaluation protocol. 

From the development of the framework, we identify issues to be addressed if it were used as a standard diagnostic auxiliary tool by domain experts and other stakeholders. Critical aspects of being mindful of such technology include data sharing and use, reproducibility, domain expertise, and value-charged choices carried out during the process.  

In summary, the contributions of our paper are threefold:
\begin{enumerate}
    \item We propose a framework for detecting gender biases in court decisions, comprising an experimental pipeline of binary classification on the presence of gender biases in court decisions issued in Brazilian Portuguese, which can be reproduced by domain experts with some technical training;
    \item We introduce two datasets of court decisions issued by the São Paulo state Court of Justice (Brazil) in gender-based violence cases, DVC (Domestic Violence Cases) and PAC (Parental Alienation Cases), with annotation (partial and complete, respectively), their metadata on a range of legal attributes, their documentation, and the description of collection, processing, and annotation~protocols;
    \item We highlight and describe critical features that should be present in computational technologies proposed as support tools for assessing court activity in gender issues, in particular, and in human rights issues, in~general. 
\end{enumerate}

The remaining of this text is organized as follows. In Sec.~\ref{sec:gender}, we explain the motivation behind addressing gender stereotypes in court decisions while presenting concepts related to gender biasing. In Sec.~\ref{sec:rel}, we briefly present part of the literature that also addresses the automatic detection of gender biases in the legal domain. In Sec.~\ref{sec:framework}, we describe the case study data and framework: the methodology followed to build them, a baseline validation protocol, and the main results observed from the experimental pipeline. In Sec.~\ref{sec:disc}, we propose a discussion based on what we identify as critical technical, legal, and ethical aspects to be addressed for this kind of technology to fulfill its purposes. In Sec.~\ref{sec:final}, we elaborate on our findings and prospects for future directions. Finally, we present an ethics statement.  

\subsection{Institutional Gender Bias}
\label{sec:gender}

Stereotyping assumes one's characteristics or roles due to belonging to a particular group; when associating a feature with a group and assuming its members share this feature, disregarding their individual traits, we are stereotyping them. Therefore, a stereotype is a generalized view or preconception about a group~\cite{Cook2010}. A gender stereotype exists when such a view is related to the gender of its target. Humans stereotype each other for many reasons: to maximize simplicity and predictability, to assign difference, to script identities --- in general, to make sense of the world by reducing its complexity~\cite{Cook2010}. Stereotypes can reflect statistical evidence about a group, and they are not necessarily negative; however, some might be noxious. 

Gender stereotypes, in particular, tend to be especially harmful towards women and represent a ``challenge in combating sexism, which is often perpetuated through stereotypes'', according to \citet{Cook2010}. The authors describe how such generalizations might help degrade women, diminish their dignity, disproportionately add to their burden, and hamper their access to rights or justified benefits. 

In that sense, illegitimate gender stereotyping is a pervasive human rights violation~\cite{Cusack2013}. The Convention on the Elimination of All Forms of Discrimination against Women (CEDAW)~\cite{cedaw}, having 189 parties as of 2024\footnote{See \href{https://treaties.un.org/Pages/ViewDetails.aspx?src=TREATY&mtdsg_no=IV-8&chapter=4&clang=_en}{this United Nations Treaty Collection page} for a complete and updated list of signatures and ratifications, accessions, or successions.}, expresses that state parties must take adequate measures to eliminate prejudices and practices based on stereotyped concepts of gender roles; authorities and institutions, including tribunals, must eradicate discrimination against women.

However, institutions themselves are often the venue in which harmful gender stereotyping occurs and unfolds into destructive consequences. Legislative processes, court rulings, and the Law itself reflect social, political, and economic relations present in society; therefore, despite their neutrality rhetoric, they frequently reinforce gender discrimination practices~\cite{Becerra2012}.  
Several studies have addressed how judicial proceedings issue gender stereotyping acts and some consequences of this~\cite{Becerra2012, PenasDefago2015, FernandezRodriguezdeLievana2015, Almeida2017, Moyses2018, camila-2021}. Particularly in Brazil, Federal Law 11340/2006 (\textit{Lei Maria da Penha})~\cite{penha} creates legal mechanisms, including proceeding rules, aiming to prevent and repress violence against women, according to guidelines provided by the country's Federal Constitution~\cite{cf} and the CEDAW. However, there is evidence that Brazilian courts often disregard some of its provisions while relying on noxious stereotypes, resulting in inappropriate institutional responses to women affected by gender violence~\cite{Almeida2017, Moyses2018}. 

Studies providing that kind of evidence are mostly based on traditional methods from the social sciences (e.g., content analysis). In general, data of interest --- usually decisions and other physical or digital documents issued by courts --- is collected manually or through web scraping. Quantitative analysis is limited to tens or no more than a few hundred documents and is performed by a person or group. In that context, natural language processing tools might help expand possibilities of analysis of such documents --- after all, language itself might contain traces of stereotyping~\cite{Luchjenbroers2007, Sap2020}. 

The protocol we describe enables the collection and extraction of patterns from a larger volume of texts since it allows the automation of processes currently handled by humans. It provides ways for legal practitioners and researchers to answer domain questions and analyze the presence and implications of gender biases in courts. It also contributes as a methodology that can be used to apply automatic text classification techniques in the social sciences. We argue, however, for a cautious approach when designing and implementing this kind of technology.    

\section{Related work}
\label{sec:rel}

Previous work has addressed the issue of automatic detection of gender biases in legal contexts, from which we stress the following ones. 

\citet{pinto} proposed a project to develop a linguistic model and a tool to perform such a task over a (manually annotated) corpus of legal sentences published by the Portuguese Ministry of Justice on gender-based violence cases. While their approach is similar to the one we propose, they have not published results or settled on a methodology. On the other hand, \citet{Sexton2020} showed results on using supervised classification models to detect gender biases in Fijian court documents issued in the context of gender violence cases. Their dataset has 13,384 court documents, of which 809 were annotated --- the same strategy we used in our framework. However, they evaluated performance on different models: a support-vector machine, convolutional neural network architectures, and BERT-based architectures. They all showed promising results, but the authors stress challenges such as managing overfitting --- due to the low availability of annotated data ---, having experiments hampered by limitations on computational processing, and dealing with data heterogeneity. There are overlaps between their results and the ones we present; they do not mention non-technical challenges or ethical constraints that might have been present.  

\citet{sevim23} reconstructed the corpora used to train Law2Vec, a legal domain-specific word embedding model, to assess gender biases present in legal documents from different sources. While their work focuses on legislation rather than court decisions, they provide technical insights for evaluating language biases in the legal domain and mention ethical aspects concerning the task --- such as the potential of unfair outcomes when informed by biased applications. 

\citet{baker21}, on its turn, proposed an approach focused on determining the presence of gender bias within the US judicial system, primarily based on case law. From a dataset of over 6.7 million decisions, the author proposes new ways for automating the creation of biases-related word lists and uses clustering algorithms to group the documents; their main contribution relies on stressing that consistent definitions of biases are essential to achieve consistent results. That conclusion is aligned with what we observed while developing our framework and the beyond-technical aspects that we identify as critical in developing such technologies, as discussed in Sec.~\ref{sec:disc} --- particularly regarding the importance of domain expertise. 

\section{Framework}
\label{sec:framework}

This section describes our methodology for developing the framework under study, the data, and the main results observed from the proposed experimental pipeline. 

\begin{figure}[htbp]
    \centering
    \includegraphics[width=\textwidth, trim={0.5cm 0.5cm 0.25cm 0.5cm},clip]{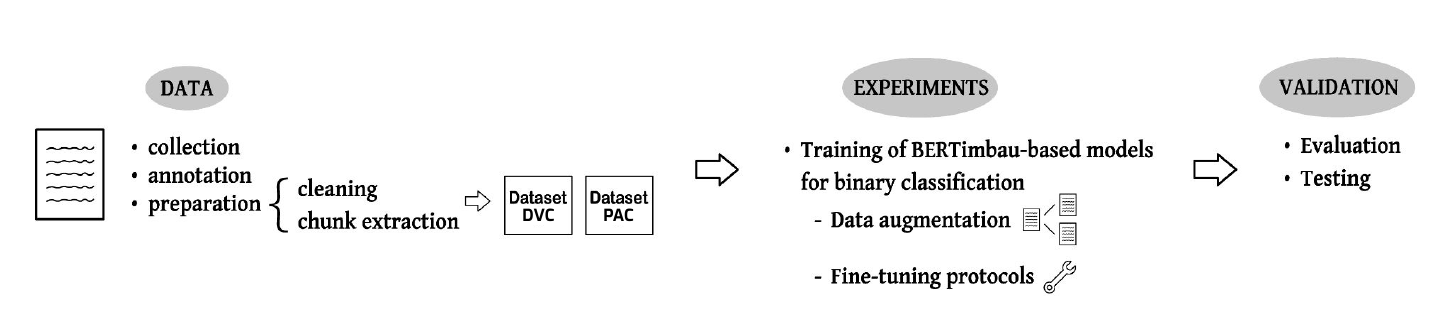}
    \caption{High-level view on the methodology. It comprises three blocks: the first one, Data, includes collection, annotation, and preparation with cleaning and chunk extraction, generating Domestic Violence Cases (DVC) and Parental Alienation Cases (PAC) datasets; they are the input of the second block, Experiments, containing training of BERTimbau-based models for binary classification, with data augmentation and fine-tuning protocols. Finally, the third block, Validation, includes evaluation and testing.}
    \Description{A diagram of blocks and arrows illustrates the methodology comprising three blocks: the first one, Data, includes collection, annotation, and preparation with cleaning and chunk extraction, generating DVC and PAC datasets; they are the input of the second block, Experiments, containing training of BERTimbau-based models for binary classification, with data augmentation and fine-tuning protocols. Finally, the third block, Validation, includes evaluation and testing.}
    \label{fig:methodology}
\end{figure}

Fig.~\ref{fig:methodology} summarizes the methodology. It starts with protocols of collection, annotation, and preparation of two datasets of Brazilian court decisions, whose texts are cleaned and transformed into chunks. This step aims to adequate the data's content and size to feed the models that classify them.

Classification is performed in the experimental phase. We ran a set of experiments over BERTimbau-based models~\cite{bertimbau}, BERT-based pre-trained models for Brazilian Portuguese, with different degrees of data augmentation, to train them to differentiate between biased and non-biased chunks of labeled text. In this phase, we applied different fine-tuning protocols over the pre-trained networks, using our own data to adjust their parameters.  

We used the training and validation sets to teach and evaluate the models. For evaluation, we used loss metrics and balanced accuracy. Besides evaluating model performance on the validation set, the validation methods include a baseline testing pipeline. While our test sets are too small to pose statistically significant validation results, we ran a pipeline that uses the best versions of the trained models to label all the texts of court decisions compatible with the framework. It could, therefore, be used in enriched versions of our datasets or new ones. 

Complete documentation of technical aspects of the framework, data, and codes can be found in the project's open repository\footnote{Available at \url{https://github.com/ra-ysa/gender_law_nlp}.}. The datasets \cite{datasets} can be downloaded and used under conditions as discussed in Sec.~\ref{sec:disc}. 

\subsection{Data}

All of the decisions used as input for our investigation were issued in the second instance of the São Paulo state Court of Justice (TJSP, \textit{Tribunal de Justiça de São Paulo}), one of the 27 Brazilian state courts (one for each of the country's federative units). Its jurisdiction reaches criminal and civil state-level disputes in virtually all but elections-, military-, and labor-related matters, which fall under the competence of special courts. 

Gender biasing in legal settings can take place in diverse ways, given the pervasiveness of gender-related stereotyping in culture and social institutions. In court, decisions in which gender stereotypes play a role as part of the motivation seem to emerge regularly in cases of domestic violence \cite{FernandezRodriguezdeLievana2015, Moyses2018}, custody and other family disputes \cite{atala, camila-2021}, health care and reproductive rights \cite{bolivia, PenasDefago2015}, and rape \cite{PenasDefago2015, Almeida2017}. Therefore, to analyze such biases on a large scale, sentences issued in these contexts often provide the content under investigation. In the Brazilian justice system, they usually fall under the jurisdiction of state common courts, such as TJSP. 

We built and performed experiments over two datasets of decisions issued by the court: DVC, which comprises 1,604~decisions issued between 2012 and 2019 in domestic violence-related cases, and PAC, which comprises 49~decisions issued between 2012 and 2019 in parental alienation-related civil and criminal cases. In both datasets, domain experts selected search criteria and instances.

Besides the data selection derived from the work of the experts, other criteria behind the choice for the state of São Paulo include: (a) data volume, given that TJSP has the highest amount of legal cases among all of the courts in the country (more than 28 million as of 2022 \cite{cnj-numeros}); (b) ease of collection, since the court's official website and search engines allow for data scraping, and auxiliary tools are available\footnote{While there are scraping tools for data produced in other courts, each website and search engine has its standard, which hampers the possibility of using other sources.}; (c) proximity and familiarity with the local court, given that the authors and collaborators of this work are all based and affiliated in the state (and some of us have previously worked with the institution). By not including data from other courts, we acknowledge that we cannot assess the protocol's performance and limits in a more diverse range of regional particularities. 

\subsubsection{Data Annotation}

Since relying on minimally supervised approaches, we identified the need to partially annotate the data for information not provided in the extraction phase. Metadata and other features, from automatic extraction to manual annotation, were added to each decision. Although most features were left out of the experimental pipeline (which focused on the biases only), they contain information that could be explored in future research. Additionally, for some attributes, categories can be clustered based on similarity to reduce the dimensionality of the domain. For DVC, we randomly selected $N$ documents for manual annotation, in which $N$ is the integer part of 10\% of the number of documents --- therefore, $N = 160$. PAC was annotated entirely, given its limited size.

Three people carried out the process of annotation: (1) the first author has a background in Law and Computer Science; (2) the second author is an expert in Law, human rights, and related gender issues; and (3) a domain researcher from the same field. Therefore, theoretical references --- mainly based on Cook and Cusack's work on gender stereotyping in legal contexts \cite{Cook2010, Cusack2013} ---, combined with previous domain expertise, provided the foundations on which annotation decisions were based. 

For each annotated document, an attribute \texttt{vies} (bias) contains the statement(s) in which some bias is identified; for model training and classification purposes, they are considered positive cases. Such identification was performed by (1), following guidelines and posterior qualitative validation from (2) and (3). To make the best out of the annotation labor --- since it was being made for identification of biases anyway ---, we also systematized further judicial, less interpretative information contained in the decisions, such as legal codes of the crimes under investigation, features of the parties, decision outcome\footnote{We note that biased language can exist in overall valid decisions with legitimate outcomes. However, detecting such biases is an essential task on its own since they taint the legitimacy of what should be an unbiased, soundly motivated institutional response. Assessing correlations between the presence of gender biases and decision outcomes was beyond the scope of this work; future research could explore such endeavors.}, and others. Such additional information was primarily annotated by (1) for DVC and (3) for PAC. A complete list of the annotated attributes and their domains, as well as a dictionary of values and descriptions of annotation protocols, can be found in \hyperref[app]{Appendix}.

\paragraph{Biases.} A core element of the data annotation process --- which determines what the models learn from the input texts --- is the definition of bias. Stereotyping is the assumption of one's characteristics or roles due to his or her belonging to a specific group; therefore, gender stereotypes take place when such assumptions are related to one's~gender\footnote{While we do not delve into definitions of gender --- which are better explained by other fields of science ---, we recognize the existence of different gender identities and expressions, which unfolds in such stereotyping taking place in diverse forms. For instance, one could be stereotyped due to their assigned gender, their gender identity, or their perceived gender.}.

There are several examples of institutional gender biases and their harmful consequences for the groups affected by them. In health care, for instance, access to legal abortion-related care can be delayed for younger and single women or women whose pregnancies resulted from violence perpetrated by someone close to them \cite{fonseca2020}. In legal systems, gender stereotypes can hamper access to proper institutional response in several ways: in cases of sexual violence, for example, the victim's behavior, personal history, and relationship with perpetrator(s) often play a role in how state agents perceive her testimony and other evidentiary elements \cite{coulouris}. 

Regarding the São Paulo state Court of Justice, for instance, qualitative investigations have shown tendencies of undervaluing victim's testimonies in cases of rape when she does not fulfill the ideal of an ``\textit{honest woman}'' \cite{almeida2018}; an analysis of more than 1,500 cases of domestic violence judged by the court between 2009 and 2018 revealed several biases to be stated by judges, prosecutors, and attorneys to determine whether the violence under analysis had been gender-motivated --- for example, physical features or the relationship of the subjects involved \cite{Moyses2018}. 

\citet{Moyses2018} stresses how the recognition of gender-based violence and discrimination should not depend on proof of intention in that sense by the perpetrator(s) but instead can be determined by results, according to the CEDAW. Therefore, a statement issued by a judge is biased if it is not based on evidence, results, or legal statutes but on his or her perception of how gender-weighted features of the subjects involved play a role in the case. Such perceptions often influence if --- and to which extent --- institutional response will be given to a victim.

Gender biases also play a role in decisions regarding family disputes. \citet{camila-2021} show how the scientifically unsound concept of parental alienation\footnote{Brazilian law defines parental alienation as ``the interference in the child's or adolescent's psychological development, perpetrated or induced by one of the birth parents, by the grandparents, or by who has authority, custody, or supervision over the minor, to repudiate a birth parent or causing damage to the establishment or preservation of the bonds between them'' (Law n.~12318/2010, article 2).} is used in court against women who report sexual abuse and other forms of family-perpetrated violence on their children. Stereotypes that play a role in such cases usually involve questioning the woman's nurturing capabilities and/or the child's behavior, often based on underlying conservative values on family and relationships.

\subsubsection{Data Preparation} 

Text to be used as input to the models went through a preparation phase that involved (a)~cleaning and (b) chunk extraction\footnote{Not to be confused with \textit{chunking} \cite{chunking}.}. The digital decisions are issued in PDF files; plain text extracted from them comes with some noisy elements. Although attention-based models do not require noise to be resolved, some of these elements, in our case, were known to be irrelevant, such as headers, electronic signatures, special characters, and some punctuation marks. They were, therefore, removed. 

Having plain, clean text corresponding to each annotated decision is still insufficient to feed the models of interest due to (a) its size and (b) its content. Attention-based networks typically require input text not larger than 512~tokens \cite{attention, bertimbau}. There are techniques to deal with longer texts, such as the Long-Document Transformer~\cite{longformer}; however, applying them to our data would be challenging to the point of going beyond the scope of the work, considering that our texts are written in Portuguese and are often even longer than the sizes accepted by such models. 

Additionally, court decisions display significant content that would likely be meaningless for automatic learning. Depending on the task for which the model is being trained, choosing specific parts of the content increases the odds of the learning happening. For instance, the biases that interest us tend to appear in the middle of the text amidst a broader argumentation context; other information, such as the verdict itself, is typically found in the first and/or last paragraphs. 

To overcome these issues, we applied a protocol of chunk extraction over the data. We define a \textit{chunk} as an excerpt from a text --- with no particular size but expected to be necessarily smaller than the whole content and ideally have a word count below 512 (also considering that tokenization might increase word count since a single word is typically unfolded in more than one token). The size of a chunk is defined by the number of sentences it contains; a \textit{sentence} is delimited by the presence of punctuation marks that suggest the completion of content (question marks, exclamation points, semicolons, or periods). 

Having annotated the data for attributes of interest, we can take advantage of knowing where each piece of information is most likely to be found, dismissing insignificant parts of the content. Therefore, in the training phase, each decision is represented by a chunk, or set of chunks, which make sense --- according to a domain expertise-related decision --- for the task being performed. 

\subsection{Experimental Design}

We developed an experimental pipeline of supervised learning for the task of binary classification over the annotated portion of each one of our datasets. The classification was performed over the bias attribute only.

\begin{figure}[htbp]
\centering
    \includegraphics[width=\textwidth]{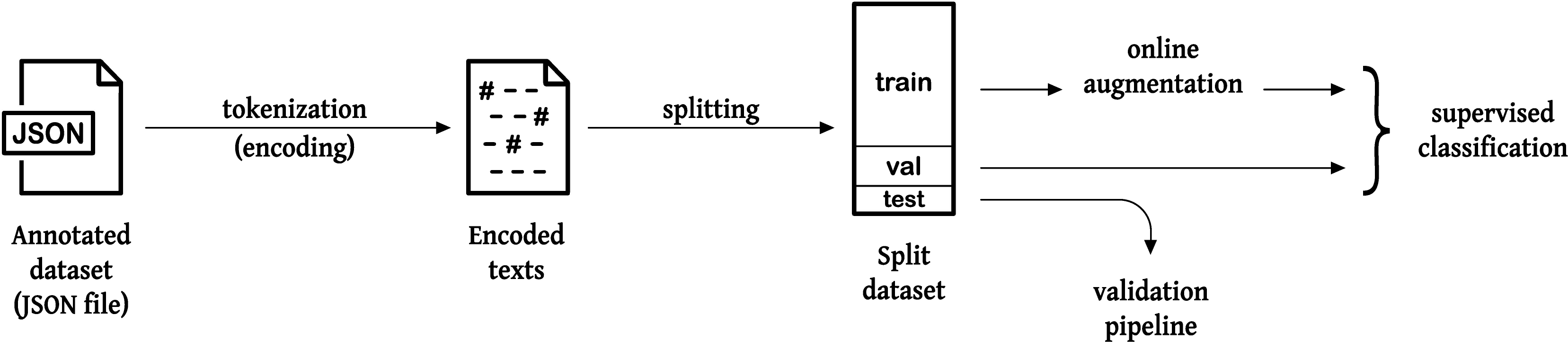}
    \caption{A representation of the experimental pipeline. It starts with a JSON file, the annotated dataset, which is tokenized (encoded). The encoded texts are split and become the split dataset, made of portions for training, validation, and testing. The training set is augmented online and, along with the validation set, is fed into a supervised classification process; the test set is fed into a validation~pipeline.}
    \Description{A diagram illustrates the experimental pipeline. It starts with a JSON file, the annotated dataset, which is tokenized (encoded). The encoded texts are split and become the split dataset, made of portions for training, validation, and testing. The training set is augmented online and, along with the validation set, is fed into a supervised classification process; the test set is fed into a validation~pipeline.}
    \label{fig:exp_pipeline}
\end{figure}

Fig.~\ref{fig:exp_pipeline} illustrates our experimental pipeline. The original annotated texts, stored in a JSON file, are encoded with the BERTimbau tokenizer; the dataset is then split in proportions of 72:18:10 for training, validating, and testing, respectively. Training and validation portions are fed into the classification model while testing instances are left for the validation pipeline.

\subsubsection{Data Augmentation}

Data augmentation, the creation of synthetic data to be used as input in automatic learning models, is a possible approach to overcome the issue of low data availability \cite{bayer2022}. It becomes then a powerful ally in our context of partial data annotation, given that augmenting data is usually cheaper than annotating it, especially when annotation is too domain-dependent, which is the case. Augmentation also partially made up for the uneven class distribution of the data: the original amount of biased decisions is around 18\% for DVC and 26\% for PAC, which were adjusted for 45\% and 212\%, respectively. 

Synthetic text can be derived from original ones through different techniques, of which we chose synonym replacement. It consists of changing a word for a synonym, thus (theoretically) preserving the original meaning and allowing the model to learn from a more diverse range of data. We performed online (during training) synonym replacement according to the following steps for each input text from the training set:

\begin{itemize}
    \item For every word of the text aside from stop words\footnote{Stop words are those with less semantic significance, usually the ones that appear frequently in text --- such as articles and prepositions. To filter them out of synonym replacement, we used the Natural Language Toolkit corpus of Portuguese stop words (\url{https://www.nltk.org/howto/portuguese_en.html}).}, we flip a coin of \texttt{weight} $= \{0, 0.3, 0.7, 1.0\}$ to decide if it will be changed for a synonym; 
    \item In case the change happens, 
    \begin{itemize}
        \item if the input text is labeled as biased, the word is replaced by (a) a synonym extracted from a domain-specific synonym dictionary \texttt{BIAS\_SYN\_DICT}, which we built from scratch based on the most bias-associated words in the annotated biased chunks, or (b) a synonym extracted from a general dictionary\footnote{We used the Brazilian Portuguese synonym dictionary from \href{https://github.com/own-pt/openWordnet-PT}{OpenWordnet-PT} \cite{coling2012}.}, in case the word to be replaced does not exist in \texttt{BIAS\_SYN\_DICT}; 
        \item otherwise, the word is replaced by a synonym extracted from a general dictionary. 
    \end{itemize}
\end{itemize}

Noticeably, there is a trade-off between the augmentation weight (expected to correlate to model learning performance) and the processing cost of the experiment.   

\subsubsection{Model and Parameters}

The binary classification task on the bias for DVC and PAC was learned by the BERTimbau model \cite{bertimbau}. While originally trained for masked-language modeling\footnote{See \url{https://github.com/neuralmind-ai/portuguese-bert}.}, the model can be used as a classifier through its Hugging Face interface\footnote{See \url{https://huggingface.co/neuralmind}.}. We imported the \texttt{bert-base-portu\-guese-cased} version of the model as an \texttt{AutoModelForSequenceClas\-sification}.

While the original BERTimbau embeddings were preserved (frozen) during learning, we fine-tuned some of the model's parameters with our inputs. For each dataset and augmentation weight, two fine-tuning protocols were used:

\begin{enumerate}
    \item Baseline protocol (\texttt{BertBaseline} class): the whole original network is preserved (frozen) except for the last layer, where the actual classifier is; 
    \item Deep fine-tuning protocol (\texttt{BertFineTuner} class): we preserve (freeze) all but the last \texttt{N\_L} $= 5$ layers of the network, over which the fine-tuning is performed. The value of \texttt{N\_L} was chosen empirically after preliminary experiments showed the optimal value to be between $4$ and $6$ since overfit increases significantly for \texttt{N\_L} $\geq 7$. Processing costs also increase prohibitively for higher \texttt{N\_L} values.
\end{enumerate}

Having two datasets, four augmentation weights, and two fine-tuning protocols, we performed 16 final training experiments. In all of them, the following parameters were used: (a) a batch size of 32 instances; (b) $20$ epochs of training; and (c) a loss-based optimization with PyTorch's \texttt{AdamW} optimizer and \texttt{CosineAnnealingLR} scheduler. 

\subsection{Evaluation and Validation Methods}

The low availability of data hampered the validation of our protocol over the test set since only 10\% of the annotated portion of each dataset was set aside for testing --- whose results, therefore, are not statistically significant in our context. However, the validation pipeline can be explored in future work with larger amounts of annotated data besides serving as a baseline tool for final users interested in using our model over full, non-annotated decisions. 

In this phase, we chose the version of the trained model for each dataset that showed the best-balanced accuracy value in the validation set over all experiments. We split the whole content of each decision into chunks; for a given decision, if any of its chunks are classified as biased by the model, all of its chunks are given the same classification. This protocol considers that, when not in the learning phase, detecting bias in one portion of a decision is equivalent to detecting the whole decision as biased. 

We used confusion matrices to help visualize model performance on the epochs with the lowest loss value. 

\subsection{Main Findings}

Our main experimental results are summarized in Tables \ref{tab:res1} and \ref{tab:res2}. They show, for each dataset and fine-tuning protocol, the best-balanced accuracy for training (label `T') and validation (label `V') sets, as well as the first epoch in which it was observed. For each dataset, we chose the trained model with the best-balanced accuracy value in the validation set to be used in the testing pipeline.  

\begin{table}
\centering
\caption{Summarized results for DVC. `T' stands for training;  `V' stands for validation.}
\label{tab:res1}
    \begin{tabular}{ccl}
    \toprule 
    Fine-tuning protocol & Augmentation  \texttt{weight} & Best-balanced accuracy (\%) (epoch) \\ \midrule 
    
    \multirow{4}{*}{Baseline} & 0 & 76.54 (T) (17), 69.15 (V) (19) \\
    
     & 0.3 & 74.92 (T) (16), 71.31 (V) (19)\\
    
     & 0.7 & 75.54 (T) (19), 73.32 (V) (16) \\
    
     & 1.0 & 72.95 (T) (19), 74.52 (V) (16)\\ \cmidrule(l){1-3}
    
    \multirow{4}{*}{Deep} & 0 & 100.00 (T) (10), 85.74 (V) (19)\\
    
     & 0.3 & 100.00 (T) (10), 88.86 (V) (7)\\
    
     & 0.7 & 100.00 (T) (13), 86.70 (V) (12) \\
    
     & 1.0 & 100.00 (T) (14), 85.74 (V) (8)\\ 
     \bottomrule
     \end{tabular}
 \end{table}

\begin{table}
\centering
\caption{Summarized results for PAC. `T' stands for training;  `V' stands for validation.}
\label{tab:res2}
    \begin{tabular}{ccl}
    \toprule 
    Fine-tuning protocol& Augmentation \texttt{weight} & Best-balanced accuracy (\%) (epoch) \\ \midrule 
    
    \multirow{4}{*}{Baseline} & 0 & 74.50 (T) (14), 83.93 (V) (19) \\
    
     & 0.3 & 73.74 (T) (16), 85.71 (V) (19)\\
    
     & 0.7 & 74.59 (T) (18), 85.71 (V) (17) \\
    
     & 1.0 & 72.47 (T) (16), 87.90 (V) (19) \\ \cline{1-3}
    
    \multirow{4}{*}{Deep} & 0 & 100.00 (T) (8), 87.90 (V) (3)\\
    
     & 0.3 & 100.00 (T) (9), 94.05 (V) (5) \\
    
     & 0.7 & 100.00 (T) (11), 94.05 (V) (9)\\
    
     & 1.0 & 100.00 (T) (11), 95.83 (V) (11)\\
     \bottomrule
    \end{tabular}
\end{table}

Data augmentation helped make up for the low availability of annotated data. In most experiments, values of balanced accuracy increase with the augmentation weight while overfitting decreases. In the deep fine-tuning protocol, an augmentation \texttt{weight} of $0.3$ increased accuracy significantly, especially in DVC. Therefore, combining this strategy with partial data annotation helps achieve a reasonable trade-off between the cost of building a quality dataset and getting good performance in the task that we want a model to learn. 

Overall, overfit is more prevalent in experiments that used the deep fine-tuning protocol over the baseline ones; they also showed better evaluation metrics and less confusion between classes. For instance, Tables \ref{tab:cm_ap_baseline} and \ref{tab:cm_ap_ft} show confusion matrices of results over the validation set of PAC at each fine-tuning protocol, using the maximum augmentation weight. While the deep fine-tuning protocol slightly increased false negatives, overall classification was more accurate, significantly decreasing false positives. 

\begin{table}[h!]
\centering
\caption{Confusion matrix for results over the validation set of PAC (\textit{baseline fine-tuning protocol}, augmentation \texttt{weight} $= 1.0$).}
\label{tab:cm_ap_baseline}
\begin{tabular}{crcc}
\toprule
 &  & \multicolumn{2}{c}{\textbf{Predicted class}}  \\ 
 &  & Non-biased (\%) & Biased (\%) \\ \cmidrule{3-4} 
\multirow{2}{*}{\textbf{Actual class}} & Non-biased & 21.05 & 15.79\\ 
 & Biased & 2.63 & 60.53 \\ \bottomrule
\end{tabular}
\end{table}

\begin{table}[h!]
\centering
\caption{Confusion matrix for results over the validation set of PAC (\textit{deep fine-tuning protocol}, augmentation \texttt{weight} $= 1.0$).}
\label{tab:cm_ap_ft}
\begin{tabular}{crcc}
\toprule
 &  & \multicolumn{2}{c}{\textbf{Predicted class}}  \\ 
 &  & Non-biased (\%) & Biased (\%) \\ \cmidrule{3-4} 
\multirow{2}{*}{\textbf{Actual class}} & Non-biased & 34.21 & 2.63\\ 
 & Biased & 5.26 & 57.89 \\ \bottomrule
\end{tabular}
\end{table}

Using an augmentation weight above zero, combined with the deep fine-tuning protocol, is the best approach regarding model performance between the ones we tested; however, in future work, it should be enhanced with strategies to mitigate overfitting. 

Although our approach makes sense from an automatic learning perspective, the lack of robust validation prevents us from assessing the generalization capabilities of the models. As discussed in Sec.~\ref{sec:future}, future directions could address this issue with larger datasets --- which could include collecting new data and/or enriching DVC and PAC with more annotated instances. Adapting the protocol to be more annotation-independent would allow for exploring other validation possibilities. 

\section{Discussion}
\label{sec:disc}

Computer-enhanced information extraction provides possibilities of automating tasks previously performed by humans, increasing the investigation scale. In the context of social institutions, they can be support tools for public policy diagnoses and decision-making, assessment of institutional activity, and social science research. 

In that sense, frameworks like the one we present can potentially fulfill roles in social change, as proposed by \citet{abebe20}. Detecting human rights violations in court decisions, such as harmful gender biases, helps measure the problem, diagnose how it manifests, and understand how we specify it. It is also an effort towards the call to ``study institutions up'', a concept previously described in the anthropology field, now reframed as a power-aware research focus in machine learning \cite{miceli-2022, barabas-2020}.

As in any technical intervention, clarifying its limits is essential. The development of our framework highlighted critical aspects about which one must be mindful when proposing computational tools to support decision-making in social settings. While some could be addressed in future directions (as discussed in Sec.~\ref{sec:future}), others are intrinsic to conceptual and experimental choices, and we argue that they should be contemplated in designing, implementing, and using such technologies. The following paragraphs are dedicated to describing them. 

\paragraph{Data sharing and reproducibility.}

Reproducibility is a critical quality of modern research \cite{goodman-2016, baker-2016, loscalzo-2012}, given its role in scientific scrutiny, fraud prevention and detection, and strengthening of research communities, which upholds the purpose of science as an endeavor of public interest. In computer science research, the gold reproducibility standard can be attained by publishing linked and executable code and data along with results, according to \citet{peng-2011}.   

In this context, data sharing and quality assessment emerge as an object of concern \cite{gebru-2021, blockeel-2007}. Data collecting, cleaning, labeling, and/or processing are often part of the experimental pipeline in machine learning research, which justifies interest in making them available for peers and stakeholders. However, the use and availability of datasets produced by social institutions can pose ethical and legal constraints that researchers and practitioners must consider. 

Court decisions, for instance, often contain sensitive personal information\footnote{According to the European General Data Protection Regulation \cite{gdpr} and similar provisions, personal data is sensitive when it concerns the racial or ethnic origin, political opinions, religious or philosophical beliefs, trade union membership, health or sex life, or personal genetic or biometric information.} on the subjects involved; while secrecy and/or anonymization is generally expected in those cases, it is not always properly performed, and documents with restricted personal data can end up publicly available --- and well-meaning researchers can be held accountable for propagating it. 

Even setting data sensitivity aside, other restrictions may apply. Some jurisdictions impose specific constraints on data use and disclosure --- for example, when it concerns minors, issues of social or public interest, private life matters, and others. Publicizing documents containing this kind of information can pose legal liability or be ethically debatable, given that it amplifies risks for the subjects involved. Those risks include violation of privacy and intimacy rights, exposure of confidential information, and exposure of any information that might jeopardize the safety or integrity of the subject(s) involved in a legal case. Such violations can be particularly harmful in human rights-related disputes, which often figure socially vulnerable groups. In gender violence claims, for instance, decisions frequently contain descriptions of family and relationship dynamics, information on the health and sex life of the parties, identification of persons (including minors) and communities, and other delicate data. 

To guarantee acceptable levels of scientific reproducibility while maintaining the informational self-determination of individuals --- an elemental dimension of their human rights ---, researchers and practitioners should comply with legal and ethical guidelines for data use and availability; they can include disclosure by demand with a deed of undertaking, anonymization, and other mitigation measures \cite{benatti2022, gergely-2021, opijnen-2017}. In this work, we chose the first option; data usage and constraints instructions can be found in the project's public repository. This structure of publicization, along with the detailed methodology description provided in the work, makes up for a fair balance between scientific reproducibility and compliance with data restriction issues. Researchers should evaluate which risks and mitigation choices might apply to their context to decide on the extent of data disclosure considering available resources, aiming at preserving scientific reproducibility while respecting ethical and legal restrictions. 

\paragraph{Domain expertise.}

Domain expertise in machine learning has been an ongoing topic of scientific interest. Several researchers have addressed discussions on the matter; while the development of data-intensive tools, such as deep neural networks and large language models, brought possibilities of reducing the need for prior knowledge to deliver solutions, some argue that human expertise remains essential in the machine learning loop \cite{liu23, kumar22}, and can enhance the quality of the results~\cite{peng23, diligenti17}. 

The importance of human expertise is particularly visible when the data available for learning is not abundant, well processed, properly documented, or overall does not meet quality, safety, and ethics standards for the task being performed --- which is often the case in real-world problems. The development of our framework, for instance, required the input of domain experts in several steps of the data construction pipeline: selection of cases, collection, annotation, cleaning, chunking, and documentation. Expertise helped us determine which groups of cases were pertinent for the task, which metadata was necessary, how to identify biased decisions, which parts of the text were relevant, and what should be registered for reproducibility by other researchers and practitioners.  

The benefits of integrating domain knowledge in technical solutions go beyond the data construction. Processes such as defining and formalizing which problems should be tackled, how they should be modeled, assessing the quality of the computer-enhanced solutions, and comparing them to the available ones can all benefit from prior human expertise. For instance, previous work on text classification from domestic violence online posts showed how domain-specific embeddings produce more informative results than generic ones \cite{subramani19}. 

We argue that protocols to support decision-making in institutional contexts should not be used without human assessment, nor should their decisions be trusted without proper human (and preferably domain-based) evaluation. When designed as a diagnostic tool, their purposes are fulfilled when combined with the knowledge and abilities provided by human experts --- especially for analyzing individual cases rather than populations of instances. Expertise can also enrich context-specific validation strategies, as part of a broader participatory design \cite{suresh-2022}. We stress, however, that the participation of domain experts in the loop should be meaningfully integrated into the process; ``participation washing'' \cite{sloane22} should be avoided.  

\paragraph{Value-charged choices.}

Values are pervasive in every scientific endeavor --- not only in pre- and post-scientific activities, such as the definition of problems and application of results but also at the core of scientific reasoning. The acceptance and rejection of hypotheses require scientists to make value judgments \cite{rudner}; scientific investigations in which errors can cause non-epistemic\footnote{Social, ethical, and political aspects are examples of non-epistemic values in science~\cite{douglas00}.} consequences require non-epistemic values to be considered in methodological choices, data characterization, and interpretation of results \cite{douglas00}. 

Particularly in machine learning research and practice, value-charged decisions play a role in different process stages, including data construction (comprising selection, processing, annotation, and availability), choice of models and parameters, and selection of evaluation metrics. When using natural language processing to address gender issues, for instance, one's views on gender --- and gender-based stereotypes, if pertinent --- can influence how these steps will be performed. 

Our definitions of gender and biases are intrinsically limited by the references we have had access to, as well as our own interpretations and perceptions of such references --- even if logical, well-based, and scrutinizable, which are the qualities that make them acceptably scientific. Building a tool to learn gender biases from court decisions requires some degree of a discretization of concepts, and we should be aware of the trade-off between discretizing concepts and acknowledging their nuances, which might be lost in the process.

\citet{Larson2017} discusses theoretical and ethical guidelines to consider when dealing with gender-related concepts in natural language processing research. In that sense, the views on gender and gender-based stereotypes imprinted in our framework are aligned with the theory of gender performativity presented by Judith Butler in 1990 and explored in related work since \cite{butler}, according to which ``language is a part of gender performativity, and (...) a key part of how we transmit and maintain stereotypes, (re)produce meaning, and navigate systems of power'' \cite{devinney-2022}. 

\section{Final Remarks}
\label{sec:final}

This work presents an attention-based natural language processing binary classification protocol to address the issue of automatic gender bias detection in Brazilian court decisions delivered in the context of gender-based violence cases. Our framework comprises:
\begin{enumerate}
    \item The collection, partial annotation, and preparation of data --- which, in our case, was extracted from the São Paulo state Court of Justice and made up of two datasets, DVC and PAC, built with the help of domain experts;
    \item The usage of an experimental pipeline based on BERTimbau, a pre-trained BERT model for the Brazilian Portuguese language;
    \item The evaluation of such pipeline and a baseline validation protocol. 
\end{enumerate}

We also described critical features concerning data sharing and use, reproducibility, domain expertise, and value-charged choices that should be considered in the design and implementation of computational technologies proposed as support tools for the assessment of court activity, especially in human rights-related issues, such as the identification of gender biases. 

Automatic detection of gender biases in court decisions allows domain experts to address some of their research inquiries and enrich diagnoses on how such harmful practice is institutionally perpetrated. The underlying hypotheses behind this project are that (a) gender biases and stereotypes can be detected in judicial decisions on a large scale, and (b) natural language processing offers suitable approaches to detect them. While there are caveats behind the answer for each one of them and the protocol we developed needs improvement, we consider our results to corroborate both hypotheses; in that sense, the model we propose can be used and understood as proof of concept. 

Data was collected automatically due to the availability of scraping tools, combined with input from domain experts --- which was crucial throughout the whole work. However, our approach has scalability issues, especially for PAC, since the tools only sometimes worked as expected and had to be adapted for our instances and complemented with manual interventions. 

Annotating our data also required domain knowledge, which hampers the possibility of annotating full large datasets --- after all, that would defeat the purpose of using automatic strategies to facilitate the human work of analyzing each decision. Still, domain knowledge remains an ally rather than an obstacle since it allowed us to build the dataset from scratch, mindfully annotate it, choose and calibrate adequate models, create a validation pipeline for the protocol, and thoroughly document and be aware of the references behind our decisions.

Overall, while our protocol has shown fair results and indicates a promising approach, we do not vouch for its indiscriminate use, especially not before improvements are made to the automatic learning process and the critical features described in Sec.~\ref{sec:disc}. The following section describes limitations that could be addressed in future endeavors.

\subsection{Future Directions}
\label{sec:future}

Although we propose a complete pipeline for data collection and automatic gender bias detection in court decisions issued in Brazilian Portuguese in gender-based violence cases, many issues remain to be addressed and could be explored in future directions. Those include: 

\begin{itemize}
    \item \textbf{Datasets:} Our approach could be applied to, validated in, and/or expanded for other datasets of court decisions featuring gender issues. Besides enhancing the scalability features of our protocol of collection, documents issued by other courts, in different time frames, or a more diverse range of cases and attributes (including the ones for which we provided annotation protocols) could be explored in that sense;
    \item \textbf{Use by domain experts:} Since our pipeline requires technical training, further work could integrate other forms of participation and improve its usability --- and, therefore, its reach power; 
    \item \textbf{Modeling:} A more diverse range of models can be explored for automatic bias detection. They might include domain-specific fine-tuned models, approaches based on feature extraction, and approaches based on traditional models rather than attention-based ones. Examining such options could improve performance results and enrich our understanding of the task; 
    \begin{itemize}
        \item \textbf{Use of other large language models:} The release of pre-trained large language models in the past months --- such as the GPT series \cite{openai-gpt4} and LLaMA \cite{llama}, as well as comparable options trained in languages other than English, such as Sabiá for Brazilian Portuguese \cite{pires-sabia} --- redefined standards for state-of-the-art performance in many natural language processing tasks. The possibilities offered by them for our investigation could be explored in future research; 
    \end{itemize}
    \item \textbf{Validation:} Validation of our protocol over the test sets was hampered by the scarcity of annotated data, causing testing results to be statistically insignificant. Therefore, although experimental results are fair and we present a usable validation pipeline, a more robust evaluation of its generalization capability remains to be developed --- yet another dimension in which more domain expertise participatory efforts should be integrated; 
    \item \textbf{Annotation:} Dependency on domain-specific annotation, which causes low annotated data availability, can be addressed differently. Annotating more data improves availability, but it is costly; data augmentation is a cheaper, feasible option, which we chose in this project. Future directions could explore automatic annotation protocols and/or unsupervised techniques to make the pipeline more annotation-independent.
\end{itemize}

\section*{Ethics Statement}
\label{sec:ethics}

The main purpose of our contributions is to provide a responsible approach for researchers and practitioners interested in investigating gender biases and related features in court decisions, particularly those issued in Brazilian Portuguese. We foresee our protocols and guidelines being helpful for them to, among others: 

\begin{itemize}
    \item Decide whether and to which extent to disclose datasets made of court documents, especially in gender-based violence and other human rights violations-related cases;
    \item Collect, process, and annotate court documents as a data source for automatic learning models by either using our protocol or deriving similar ones; 
    \item Explore the information provided by our datasets to investigate institutional gender biases in Brazilian courts, especially from the state of São Paulo, as well as other features associated with the metadata and annotation we provided; 
    \item Use, expand, and assess our experimental pipeline and baseline testing protocol to detect gender biases in court decisions on a large scale, thus unlocking helpful diagnostic information on the matter. 
\end{itemize}

Despite the positive impacts that our work might induce, we must acknowledge that distorted and/or unpredicted interpretations and uses derived from it can arise, which could lead to unwanted outcomes. These include but are not limited to:

\begin{itemize}
    \item Breach of the terms of the deed of undertaking to which one must abide to access our datasets --- which, although entails liability, carries the risks associated with wrongfully using and/or disclosing their content; 
    \item Bypassing human assessment and previous domain-informed knowledge when using and evaluating our tools and their derived results could lead to misdiagnosis of the issues we propose to address. Examples include:
    \begin{itemize}
        \item dismissing other sources of institutional gender biases in justice systems; 
        \item wrongfully pointing specific individuals or court chambers as bias perpretrators; 
        \item over or underestimating occurrences of institutional gender biases in Brazilian courts.
    \end{itemize}
\end{itemize}

We try to mitigate unwelcome derivations of our work by thoroughly describing its processes, methods, caveats, and intended implications, also believing that foreseeing associated risks within reason helps us understand the limits and possibilities offered by our approach. 

\begin{acks}
This work was funded by CAPES/Brazil and FAEPEX/Unicamp, having been developed mostly at the Recod.ai lab in the University of Campinas, Brazil. The Recod.ai lab is supported by projects from FAPESP/São Paulo, CNPq/Brazil, and CAPES.
R.~Benatti is currently funded by the Deutsche Forschungsgemeinschaft (DFG, German Research Foundation) under Germany’s Excellence Strategy --- EXC number 2064/1 --- Project number 390727645.
F. Severi is supported by the University of São Paulo's Law School of Ribeirão Preto. 
S. Avila is partially funded by CNPq PQ-2 grant 316489/2023-9, FAPESP 2013/08293-7, 2020/09838-0, 2023/12086-9, and H.IAAC (Artificial Intelligence and Cognitive Architectures Hub). 
E. Colombini is partially funded by CNPq PQ-2 grant 315468/2021-1 and H.IAAC. 
The authors thank the International Max Planck Research School for Intelligent Systems (IMPRS-IS) for supporting Raysa Benatti. We also thank Camila Maria de Lima Villarroel, Luanna Tomaz de Souza, Rodrigo Frassetto Nogueira, and Konstantin Genin for helpful discussions and feedback. Finally, we are grateful for the reviewers of this work, who enriched it with their evaluation and comments.
\end{acks}

\bibliographystyle{ACM-Reference-Format}
\bibliography{sample-base}

\appendix
\label{app}

\section{DVC Dataset: Domestic Violence Cases}

Table \ref{table:annotate_lesao} summarizes the annotated attributes and their domains, followed by a dictionary of values and descriptions of annotation protocols.

\begin{table}[hp]
\centering
\caption{Data attributes annotated to 10\% of the documents in DVC.}
\label{table:annotate_lesao}
\begin{tabular}{p{3.3cm}p{6cm}p{5.5cm}}
\toprule
Attribute name & Description & Domain\textsuperscript{(a)} \\ 
\midrule
\texttt{apelante} & identification of the appellant party (anonymized if natural person) & Any combination of name initials; \texttt{mpsp} \\
\texttt{apelante\_genero} & gender of the appellant & \texttt{masc; fem;} \texttt{masc\_trans; fem\_trans} \\
\texttt{apelado} & identification of the appealed party (anonymized if natural person) & Same as \texttt{apelante} \\
\texttt{crime} & legal code(s) of crime(s) under analysis in the case & \texttt{cp129p6; cp129p9; cp147;} \texttt{cp150p1; cp330; cp331;} \texttt{cp345; ct306; lcp21; lcp65}\\
\texttt{vitima} & victim(s) main relationship with the defendant & \texttt{comp; esposa; namo;} \texttt{ex; fam\_ex; rel\_ex;} \texttt{filha; ent;} \texttt{irma; irmao; sob; cnh;} \texttt{mae; pai; tia; amiga} \\
\texttt{vitima\_genero} & gender of the victim(s) & Same as \texttt{apelante\_genero}\\
\texttt{pena\_original} & time of prison punishment, in months, issued against the defendant in first instance & $[0, 23.5]$ \\
\texttt{requer} & main request(s) made by the appellant & \texttt{abs; cond; abrand;} \texttt{desclass; cond\_sem\_qual;} \texttt{afast\_altern; maj; conc\_mat} \\
\texttt{requer\_subsid} & subsidiary request(s) made by the appellant & \texttt{abrand; desclass; afast\_sursis} \\
\texttt{requer\_motivo} & main reason(s) claimed by the appellant & \texttt{provas; aut\_mater;} \texttt{insig; atip; aus\_dolo;} \texttt{leg\_def; conf; cp129p4;} \texttt{inimputab; } \texttt{fato; jur; vit;} \texttt{antec; n\_antec} \\ 
\texttt{mp\_pj} & position stated by the Public Prosecutor's Office & \texttt{s; n; parcial; prej}\\
\texttt{resultado} & final decision on the merits\textsuperscript{(b)} of the appeal & \texttt{s; n; parcial} \\
\texttt{resultado\_razoes} & main reason(s) stated by the court to motivate the result & 
\texttt{provas; aut\_mater; fund\_legal;} \texttt{bis\_in\_idem; jur;} \texttt{vit; conf; n\_antec;} \texttt{imputab; leg\_def; circ; presc;} \texttt{prej}\\
\texttt{pena\_atual} & penalty issued against the defendant after the appeal & $[0, 15.17]$; 
\texttt{idem; sursis; sem\_sursis;} \texttt{abrand\_reg; sem\_serv; prej} \\ 
\texttt{vies} & biased statement(s) identified in the decision & See section on biases \\
\texttt{vies\_alvo} & target(s) of the biased statement(s) & \texttt{vitima; reu; test;} \texttt{abs\_mul; abs\_reu; soc} \\
\bottomrule
\end{tabular}
\footnotesize
\flushleft
(a) An empty value is part of the domain for all the attributes. It was omitted from the table to avoid redundancy. \\
(b) Discussions on appeal admissibility and other preliminary issues were not considered, except when they motivated acquittal (e.g., in case of statute of limitations).
\end{table}

\subsection{Dictionary of attributes}
\begin{itemize}

    \item Gender:
    \begin{itemize}
        \item \texttt{masc}, \texttt{fem}, \texttt{masc\_trans}, and \texttt{fem\_trans} mean, respectively, cisgender masculine, cisgender feminine, transgender masculine, and transgender feminine. While we acknowledge the existence of other genders, their labels are not used in official court records to the best of our knowledge. We assigned gender labels considering: (a) the usual gender attributed to the name of the subject; (b) pronouns used in the decision to refer to the subject; (c) gender descriptions stated in the document. Gender self-identification would have been a primary criterion if stated in the documents, which is not the case. \\
    \end{itemize}
    
    \item Appellant / Appealed parties: 
    \begin{itemize}
        \item In most of the documents, \texttt{mpsp} (\textit{Ministério Público do Estado de São Paulo} --- state of São Paulo Prosecutor's Office) is the appealed party since, in domestic violence cases, it is the plaintiff by default, and court decisions tend to accept its claims. The appellant is usually the person accused of the crime --- and convicted in the first instance ---, here identified by initials only. Sometimes, the opposite happens, and the prosecutor appeals against the defendant (e.g., when the first instance grants acquittal); in that case, we use the initials of the appealed person's name in the \texttt{apelado} field, and \texttt{mpsp} as \texttt{apelante}. Very rarely, the court addresses appeals from both the defendant and the prosecutor in a single decision; in that case, we annotate both parties as \texttt{apelante} and \texttt{apelado}, but the other attributes are labeled considering the defendant's appeal only. \\
    \end{itemize}
    
    \item Crime:
    \begin{itemize}
        \item \texttt{cp129p6}: unintentional bodily injury (Criminal Code, article 129, paragraph 6);
        \item \texttt{cp129p9}: intentional bodily injury perpetrated in the context of domestic relationships (Criminal Code, article 129, paragraph 9); 
        \item \texttt{cp147}: intimidation (Criminal Code, article 147); 
        \item \texttt{cp150p1}: aggravated trespassing (Criminal Code, article 150, paragraph 1); 
        \item \texttt{cp330}: defiance of the lawful authority of public servants (Criminal Code, article 330); 
        \item \texttt{cp331}: contempt of the work of public servants (Criminal Code, article 331); 
        \item \texttt{cp345}: taking the law into one's own hands (Criminal Code, article 345); 
        \item \texttt{ct306}: driving under the influence (Traffic Code, article 306);
        \item \texttt{lcp21}: assault (Misdemeanors Act, article 21);
        \item \texttt{lcp65}: harassment (Misdemeanors Act, article 65\footnote{This article was revoked in 2021 since a new related definition was included in the Criminal Code (stalking, article 147-A); however, it was valid when the facts brought to court and figuring in our dataset happened.}). \\
    \end{itemize}
    
    \item Victim:
    \begin{itemize}
        \item \texttt{comp}: partner (\textit{companheira(o)}, sometimes \textit{amásia(o)}); 
        \item \texttt{esposa}: wife; 
        \item \texttt{namo}: girlfriend or boyfriend (\textit{namorada(o)});  
        \item \texttt{ex}: ex-partner, ex-wife/husband, or ex-girlfriend/boyfriend; 
        \item \texttt{fam\_ex}: someone belonging to the ex's family;  
        \item \texttt{rel\_ex}: someone related to the ex by bonds other than family (e.g., friend or current partner); 
        \item \texttt{filha}: daughter; 
        \item \texttt{ent}: stepdaughter or stepson (\textit{enteada(o)});
        \item \texttt{irma}: sister; 
        \item \texttt{irmao}: brother; 
        \item \texttt{sob}: niece or nephew (\textit{sobrinha(o)});  
        \item \texttt{cnh}: sister-in-law or brother-in-law (\textit{cunhada(o)});
        \item \texttt{mae}: mother;
        \item \texttt{pai}: father;
        \item \texttt{tia}: aunt; 
        \item \texttt{amiga}: female friend. \\
    \end{itemize}
    
    Descriptions of both female and masculine genders were included when either (a) the abbreviation chosen for labeling the category allows for any gender to be included or (b) a case with a male victim of that category appeared in the dataset. We note, however, that the majority of victims are women. 

    Relationship status is always stated as it was when the facts happened. When the document provides conflicting information on the relationship between the victim(s) and defendant, we annotate it as informed by the victim(s); if s/he provided conflicting testimonials in different phases of the case, we interpreted the available information and circumstances to decide on a label. If the victim and defendant were legally married but factually separated, we label this attribute as \texttt{ex}. If the victim and defendant have a non-clarified companionship bond, the default label is~\texttt{comp}. \\  

    \item Penalty:
    \begin{itemize}
        \item If annotated with a number, the attributes \texttt{pena\_original} and \texttt{pena\_atual} state for how long, in months, the punishment of liberty restraint is imposed to last. Decimal parts are computed considering a 30-day month. We do not differentiate between types of prison/jail, nor annotate conditions of imprisonment and other penalties that might have been imposed, such as fines. An amount of zero means acquittal. The upper limit of the domain is established according to the longest penalty found in the annotated dataset, even if the crime under analysis can entail a longer prison time. \\        
        Penalty issued after the appeal (\texttt{pena\_atual}) can have the same imprisonment length as the original but softened by other conditions, which justifies adding information in that attribute. Its domain of textual labels is: 
        \item \texttt{idem}: same imprisonment length as first instance; 
        \item \texttt{sursis}: grant of \textit{sursis} (suspended sentence);
        \item \texttt{sem\_sursis}: dismissal of \textit{sursis}; 
        \item \texttt{abrand\_reg}: some form of mitigation of penalty other than length (\textit{abrandamento de regime});
        \item \texttt{sem\_serv}: dismissal or mitigation of community service order (\textit{sem prestação de serviços à comunidade}). \\
    \end{itemize}
    
    \item Requests:
    \begin{itemize}
        \item \texttt{abs}: acquittal (\textit{absolvição});
        \item \texttt{cond}: conviction (\textit{condenação});
        \item \texttt{abrand}: some form of mitigation of penalty (\textit{abrandamento});
        \item \texttt{desclass}: criminal downgrading to a less severe offense (\textit{desclassificação});
        \item \texttt{cond\_sem\_agr}: conviction without the aggravation motive stated in the Criminal Code, article 61 IIf\footnote{This article states the aggravation of the punishment to any crime if it is perpetrated (a) under an abuse of authority, or (b) in the context of domestic relationships --- if those circumstances are not already stated in the description of the crime itself.} (\textit{condenação sem agravante}); 
        \item \texttt{afast\_altern}: dismissal of alternative punishment (\textit{afastamento de pena alternativa});
        \item \texttt{maj}: increase of punishment time (\textit{majoração});
        \item \texttt{conc\_mat}: admission of charge stacking (\textit{concurso material});
        \item \texttt{afast\_sursis}: dismissal of \textit{sursis} (\textit{afastamento de sursis}). \\
    \end{itemize}
    
    \item Reasoning:
    \begin{itemize}
        \item \texttt{provas}: evidence; this label is used to state an argument of absence, insufficiency, or any inadequacy of evidence to support a conviction; 
        \item \texttt{aut\_mater}: used if attribution and materiality of the crime are well established (\textit{autoria e materialidade});
        \item \texttt{insig}: criminal pettiness (\textit{insignificância});
        \item \texttt{atip}: used to argue that whatever happened cannot be defined as a criminal action (\textit{atipicidade});
        \item \texttt{aus\_dolo}: absence of intention (\textit{ausência de dolo}); 
        \item \texttt{leg\_def}: lawful self-defense (\textit{legítima defesa});
        \item \texttt{conf}: confession; admission of guilt (\textit{confissão});
        \item \texttt{cp129p4}: the existence of moral motivations behind the crime or intense emotions of the perpetrator following unjust provocation made by the victim, as stated in Criminal Code, article 129, paragraph 4; 
        \item \texttt{inimputab}: unimputability (\textit{inimputabilidade});
        \item \texttt{imputab}: imputability (\textit{imputabilidade}); 
        \item \texttt{n\_antec}: absence of criminal records (\textit{não antecedentes});
        \item \texttt{antec}: presence of criminal records (\textit{antecedentes}); 
        \item \texttt{fato}: fact, i.e., anything related to factual elements of the case; 
        \item \texttt{vit}: victim (\textit{vítima}), i.e., any argument related to a deed from the victim at some point during the legal procedures (e.g., retraction of allegations); 
        \item \texttt{fund\_legal}: legal ground (\textit{fundamento legal}), i.e. anything directly linked to a legal statement; 
        \item \texttt{bis\_in\_idem}: double jeopardy; 
        \item \texttt{jur}: analogous to \texttt{fund\_legal}, but linked to a court precedent instead (\textit{jurisprudência}); 
        \item \texttt{circ}: circumstances (\textit{circunstâncias}), unspecifically; 
        \item \texttt{presc}: statute of limitations (\textit{prescrição}). \\
    \end{itemize}
    
    \item Prosecutor's position (\texttt{mp\_pj}): 
    \begin{itemize}
        \item The Prosecutor’s Office is granted the right to provide an opinion in some court cases as \textit{custos legis} (warden of the law). Such a right derives from an interpretation of its constitutional definition as guardian of social interest (Federal Constitution, article 127); there is no explicit legal provision behind it. In fact, some argue that such a deed would be unconstitutional under certain conditions since the prosecution is an interested party in many cases. Regardless, having this statement given in court is common practice, and the attribute \texttt{mp\_pj} represents its content: \texttt{s} if in favor of the appeal (\textit{sim}), \texttt{n} if against it (\textit{não}), and \texttt{parcial} if partially in favor. The same labels are used to state the final decision (attribute \texttt{resultado} (result)).
        
        \item Rarely, the first instance prosecutor (\texttt{mp} --- \textit{Ministério Público}\footnote{\textit{Ministério Público} is the prosecution institution as a whole, but, in this context, refers to the first instance division. In Brazil, generally, \textit{promotor de justiça} is the first instance prosecutor and \textit{procurador de justiça} is the second instance prosecutor. Both of them belong to the (in our case, state level) Prosecutor's Office (\textit{Ministério Público}), but when \textit{Ministério Público} and \textit{Procuradoria de Justiça} are used as distinct elements, the former refers to the first instance and the latter to the second instance divisions.}) and the second instance prosecutor (\texttt{pj} --- \textit{Procuradoria de Justiça}) state two distinct opinions; in that case, they were both annotated in the same field. \\
    \end{itemize}
    
    \item Extra considerations:
    \begin{itemize}
        \item The label \texttt{prej} is used when the analysis for an attribute was impaired (\textit{prejudicada}) due to limitations from the case itself; 
        \item Empty values were used when the corresponding attribute does not exist in the case (e.g., when prosecution appeals, it is common to omit their reasoning from the decision report since it usually repeats the arguments from the original petition); 
        \item While this dataset consists mostly of court answers to strict sense appeals (i.e., on the merits), six out of the 160 annotated documents answer to an appeal on formal and/or preliminary issues (\textit{embargos}). In those cases, all attributes were left empty since such procedural matters are beyond our scope; 
        \item All decisions described here result from a trade-off between precision and simplicity of the annotation; different contexts of use might entail different degrees of annotation diversity. We also acknowledge that the annotation process carries intrinsic biases from the researches, which we try to mitigate by (a) describing such process thoroughly, and (b) using domain knowledge as a reference behind each decision. 
    \end{itemize}
\end{itemize}

\section{PAC Dataset: Parental Alienation Cases}

Table \ref{table:annotate_ap} summarizes the annotated attributes and their domains, followed by a dictionary of values and descriptions of annotation protocols.

\begin{table}[hp]
\centering
\caption{Data attributes annotated to the documents in PAC.}
\label{table:annotate_ap}
\begin{tabular}{lp{6.5cm}p{5.25cm}}
\toprule
Attribute name & Description & Domain\textsuperscript{(a)} \\ 
\midrule
\texttt{processo} & legal case number & Any number in the format xxxxxxx-xx.xxxx.8.26.xxxx\\ 
\texttt{relator} & judge-rapporteur & Any judge assigned to operate in TJSP at second instance level\\
\texttt{orgao\_julgador} & issuing body & Any second instance court body belonging to TJSP \\
\texttt{data\_julgamento} & decision date & Any date in the format yyyy-mm-dd\\
\texttt{tipo\_recurso} & type of appeal & See dictionary \\
\texttt{colegialidade} & collegiality degree under which the decision was issued & \texttt{acordao} (at least three judges); \texttt{decisao\_monocratica} (one judge) \\
\texttt{inteiro\_teor} & availability of decision's full content & \texttt{available}\textsuperscript{(b)} \\
\texttt{assunto} & theme & See dictionary \\
\texttt{alegou\_ap} & who claimed parental alienation & See dictionary \\
\texttt{acusado\_ap} & who was accused of parental alienation & See dictionary \\ 
\texttt{viol\_mulher} & claim(s) of violence against woman & See dictionary \\
\texttt{viol\_menor} & claim(s) of violence against minor & See dictionary \\
\texttt{acusado\_viol} & who was accused of violence against minor & See dictionary \\
\texttt{resultado\_viol} & result on violence allegations & \texttt{sim} (yes); \texttt{nao} (no); \texttt{indicios} (signs) \\
\texttt{prova\_viol} & evidence used to decide on claims of violence & See dictionary \\
\texttt{resultado\_ap} & result on parental alienation allegations & See dictionary \\
\texttt{prova\_ap} & evidence used to decide on claims of parental alienation & See dictionary \\
\texttt{vies} & biased statement(s) identified in the decision & See section on biases; also includes \texttt{prej}\textsuperscript{(c)}\\
\texttt{vies\_alvo} & target(s) of the biased statement(s) & \texttt{vitima; mae; mul; soc; abs\_mul; abs\_reu; abs\_cri; prej}\textsuperscript{(c)}\\
\bottomrule
\end{tabular}
      \footnotesize
      \flushleft
      (a) An empty value is part of the domain for all the attributes. It was omitted from the table to avoid redundancy; \\ 
      (b) Originally, the contents of all selected second instance decisions from TJSP were available, and we did not change annotation made by experts unless when explicitly stated --- which is why the domain for this attribute in our dataset has only one value; \\
      (c) The entry \texttt{prej} was used when a PDF file for the decision was unavailable, preventing proper assessment of biases. 
\end{table}

\subsection{Dictionary of attributes}
\label{sec:apdict}

Since annotation for PAC was previously made by one of the experts in the context of another work (except for bias-related attributes), the domain of each attribute is more detailed, exhaustive, and redundant than in DVC. We kept the original annotation but stress the recommendation for gathering similar values depending on the context of use. \\

\begin{itemize}
    \item \texttt{tipo\_recurso}:
    \begin{itemize}
        \item Criminal merit appeals: \texttt{apelacao\_criminal}, \texttt{habeas\_corpus\_criminal}\footnote{In Brazilian legal system, the \textit{habeas corpus} is not an appeal but rather a cause per se; detailing such a technicality, however, is beyond the scope of this work.};
        \item Civil merit appeals: \texttt{apelacao\_civel}, \texttt{agravo\_de\_instrumento};
        \item Criminal appeals on procedural and/or formal issues: \texttt{embargos\_de\_declara\-cao\_criminal}, \\\texttt{recurso\_em\_sentido\_estrito}, \texttt{carta\_testemunhavel};
        \item Civil appeals on procedural and/or formal issues: \texttt{embargos\_de\_declaracao\-\_civel}, \\\texttt{embargos\_infringentes}, \texttt{embargos\_infringentes\_e\_de\_nulidade},  \texttt{agravo\_regimental\_civel}. \\ 
    \end{itemize}
    
    \item \texttt{assunto}:
    \begin{itemize}
        \item (\texttt{acao\_de\_}) (case regarding): \texttt{atentado\_ao\_pudor}: assault; \texttt{visita}: visitation; \texttt{violencia\_domestica}: domestic violence; \texttt{estupro}: rape; \texttt{guarda}: custody; \texttt{dissolucao}: dissolution; 
        \texttt{danos\_morais}: non-material damages; \\ \texttt{suprimento\_de\_consentimento}: consent supply; \texttt{guarda\_e\_visita}: custody and visitation; \\ \texttt{alimentos\_e\_dissolucao}: alimony and dissolution; \\ \texttt{alienacao\_parental}: parental alienation; \texttt{divorcio}: divorce; \texttt{ameaca}: menacing; \texttt{maus\_tratos}: maltreatment; \texttt{destituicao\_do\_poder\_familiar}: loss of parental authority; \texttt{doacao}: donation; \texttt{alimentos\_e\_guarda}: alimony and custody; \texttt{busca\_e\_apreensao}: search and seizure; \\ \texttt{danos\_morais\_e\_materiais}: material and non-material damages. \\ 
    \end{itemize}
    
    \item \texttt{alegou\_ap}:
    \begin{itemize}
        \item \texttt{genitor}: birth father; \texttt{genitora}: birth mother; \\ \texttt{ex-companheiro\_pai\_que\_nao\_e\_genitor}: former partner / non-birth father; \texttt{ambos}: both. \\
    \end{itemize}
        
    \item \texttt{acusado\_ap}:
    \begin{itemize}
        \item \texttt{genitor}: birth father; \texttt{genitora}: birth mother; \texttt{ambos}: both; \texttt{agravada}: appealed party; \\ \texttt{perita}: (female) court expert; \texttt{avo\_materna}: maternal grandmother; \\
        \texttt{avos\_paternos}: paternal grandparents; \\ 
        \texttt{atual\_companheiro\_da\_ge\-nitora}: current birth mother's partner; \\
        \texttt{genitora\_e\_sogra}: birth mother and mother-in-law. \\
    \end{itemize}
    
    \item \texttt{viol\_mulher}:
    \begin{itemize}
        \item \texttt{agressao}: physical offense; \texttt{lesao\_corporal}: bodily injury; \\ \texttt{existencia\_de\_medida\_protetiva}: presence of restraining order; \\ \texttt{ameaca\_e\_agressao}: menacing and physical offense. \\
    \end{itemize}
    
    \item \texttt{viol\_menor}:
    \begin{itemize}
        \item \texttt{abuso\_sexual}: sexual abuse; \texttt{ameaca\_e\_abuso\_sexual}: menacing and sexual abuse; \\ \texttt{maus\_tratos\_e\_abuso\_sexual}: maltreatment and sexual abuse; \\ \texttt{acusacao\_anterior\_de\_abuso\_sexual}: former complaint of sexual abuse; \\ \texttt{lesao\_corporal}: bodily injury; \texttt{agressao}: physical offense. \\ 
    \end{itemize}
    
    \item \texttt{acusado\_viol}:
    \begin{itemize}
        \item \texttt{genitor}: birth father; \texttt{madrasta}: stepmother; \texttt{companheiro\_da\_genitora}: birth mother's partner; \\ \texttt{ex-companheiro\_da\_genitora}: former birth mother's partner; \\ \texttt{companheira\_do\_genitor}: birth father's partner; \texttt{pai\_adotivo}: adoptive father; \\ \texttt{filho\_da\_companheira\_do\_genitor}: birth father's partner's son; \\ \texttt{rapazes\_que\_moram\_com\_a\_genitora}: men who live with the birth mother; \\ \texttt{esposo\_da\_avo\_materna\_e\_pai\_da\_genitora}: maternal grandmother's husband and birth mother's father; \texttt{ambos}: both. \\ 
    \end{itemize}
    
    \item \texttt{prova\_viol}:
    \begin{itemize}
        \item \texttt{in\_dubio\_pro\_reo}: in dubio pro reo; \texttt{estudo\_psicossocial}: psychosocial assessment; \\ \texttt{exame\_iml}: forensic exam; \texttt{pericia}: expert examination; \\ \texttt{estudo\_psicologico}: psychological assessment; \texttt{exame}: exam; \\ \texttt{necessidade\_de\_instrucao\_probatoria}: evidence collection needed; \\ \texttt{arquivamento\_do\_inquerito\_policial}: criminal investigation shelved; \\ \texttt{rejeicao\_da\_denuncia}: complaint rejected; \texttt{processo\_penal\_arquivado}: cri\-minal procedure shelved; \\ \texttt{nao\_houve\_oferecimento\_da\_denuncia}: complaint not presented; \\ \texttt{condenacao\_criminal}: criminal conviction; \texttt{conselho\_tutelar}: child protection services. \\
    \end{itemize}
    
    \item \texttt{resultado\_ap}:
    \begin{itemize}
        \item \texttt{alienacao\_parental\_evidenciada}: evidence of parental alienation; \\ \texttt{sindrome\_da\_alienacao\_parental\_evidenciada}: evidence of parental alienation syndrome; \\ \texttt{nao\_ocorrencia}: no parental alienation; \texttt{nao\_ocorrencia\-\_sindrome}: no parental alienation syndrome; \\ \texttt{indicios\_de\_alienacao\_pa\-rental}: signs of parental alienation; \\ \texttt{necessidade\_de\_instrucao\_probatoria}: evidence collection needed; \\ \texttt{materia\_estranha\_ao\_processo}: non-pertinent issue; \\ \texttt{existencia\_de\_acao\_declaratoria\_de\_alienacao\_parental}: parental alienation formerly acknowledged; \texttt{citacao\_de\_jurisprudencia\_pelo\_tribunal}: court mentioned precedents. \\
    \end{itemize}
    
    \item \texttt{prova\_ap}:
    \begin{itemize}
        \item \texttt{estudo\_psicossocial}: psychosocial assessment; \texttt{estudo\_psicologico}: psychological assessment; \texttt{pericia}: expert examination; \texttt{prova\_emprestada}: evidence from another case; \texttt{em\_outro\_processo}: idem.
    \end{itemize}
\end{itemize}

\section{Biases}
\label{biases}

For DVC (domestic violence cases), biased statements include:

\begin{itemize}
    \item Statements on the \textbf{relationship dynamics} between victim(s) and alleged perpetrator(s). Examples: stressing that aggression was mutual; stressing that the victim went back to, or did not break up with, the perpetrator; describing the relationship as ``troubled''; stressing that the aggression was an isolated incident in the context of the relationship; 
    \item Statements on individual gender-weighted features of the \textbf{victim} or another \textbf{woman} featured in the case. Examples: understanding that the victim's behavior gave cause to the aggression; diminishing the woman's testimony; 
    \item Statements on individual features of the alleged \textbf{aggressor}. Examples: describing the defendant's personality as either ``moderate'' or ``twisted'' and ``prone to crime''. While these stereotypes are not gender-weighted per se, they reveal a tendency to address the violence claims when the defendant is perceived as a dangerous person, and dismiss them otherwise; 
    \item \textbf{General} statements on legally and/or scientifically unsound conservative values, gender perceptions, and/or the victimhood of women in domestic violence cases. Examples: arguing for preserving the family and protecting ``societal values''; claiming women's fragility as a natural feature; deriding on women's fear of reporting their aggressors. \\
\end{itemize}

For PAC (parental alienation cases), biased statements include:

\begin{itemize}
    \item Statements on the \textbf{relationship dynamics} between mother and the alleged perpetrator. Examples: describing the relationship as ``troubled''; stressing that claims of aggression were mutual; 
    \item Statements on individual gender-weighted features of the \textbf{mother}. Examples: describing the woman as ``prone to emotional outbursts'', ``egoistic'', ``self-centered'', ``arrogant'', or ``unarticulated''; 
    \item Statements on individual features of the alleged \textbf{aggressor}. Examples: describing the defendant's reputation as ``unblemished'' or ``prestigious''; describing the defendant as a ``good father''; stressing the positive perceptions of the defendant's community on his personality and behavior;
    \item \textbf{General} statements on legally and/or scientifically unsound conservative values, gender perceptions, and/or the child's behavior. Examples: arguing in favor of traditional family settings for proper children's development; diminishing statements expressed by the child; assuming what an expected ``abused child behavior'' would look like. \\
\end{itemize}

We also annotated the target of each biased sentence. While this attribute was not used in our pipeline, it can be helpful in future work. Those include:

\begin{itemize}
    \item \texttt{vitima}: victim;
    \item \texttt{reu}: defendant; 
    \item \texttt{test}: witness; 
    \item \texttt{mae}: mother;
    \item \texttt{mul}: woman (individually --- some specific woman that does not fall under previous categories);
    \item \texttt{abs\_mul}: the collectivity of women;
    \item \texttt{abs\_reu}: the collectivity of defendants;
    \item \texttt{abs\_cri}: the collectivity of children;
    \item \texttt{soc}: society as a whole, abstractly. 
\end{itemize}
    
\end{document}